# Raw Waveform-based Speech Enhancement by Fully Convolutional Networks

Szu-Wei Fu *‡, Yu Tsao*, Xugang Lu† and Hisashi Kawai†
* Research Center for Information Technology Innovation, Academia Sinica, Taipei, Taiwan
E-mail: {jasonfu, yu.tsao}@citi.sinica.edu.tw
‡Department of Computer Science and Information Engineering, National Taiwan University,
Taipei, Taiwan
† National Institute of Information and Communications Technology, Kyoto, Japan
E-mail: {xugang.lu, hisashi.kawai}@nict.go.jp

*Abstract*— **This study proposes a fully convolutional network (FCN) model for raw waveform-based speech enhancement. The proposed system performs speech enhancement in an end-to-end (i.e., waveform-in and waveform-out) manner, which differs from most existing denoising methods that process the magnitude spectrum (e.g., log power spectrum (LPS)) only. Because the fully connected layers, which are involved in deep neural networks (DNN) and convolutional neural net-works (CNN), may not accurately characterize the local in-formation of speech signals, particularly with high frequency components, we employed fully convolutional layers to model the waveform. More specifically, FCN consists of only convolutional layers and thus the local temporal structures of speech signals can be efficiently and effectively preserved with relatively few weights. Experimental results show that DNN- and CNN-based models have limited capability to restore high frequency components of waveforms, thus leading to decreased intelligibility of enhanced speech. By contrast, the proposed FCN model can not only effectively recover the waveforms but also outperform the LPS-based DNN baseline in terms of short-time objective intelligibility (STOI) and perceptual evaluation of speech quality (PESQ). In addition, the number of model parameters in FCN is approximately only 0.2% com-pared with that in both DNN and CNN.**

## I. Introduction

Speech enhancement (SE) has been widely used as a preprocessor in speech-related applications such as speech coding [1], hearing aids [2, 3], automatic speech recognition (ASR) [4], and cochlea implants [5, 6]. In the past, various SE approaches have been developed. Notable examples include spectral subtraction [7], minimum-mean-square-error (MMSE) -based spectral amplitude estimator [8], Wiener filtering [9], and non-negative matrix factorization (NMF) [10]. Recently, deep denoising autoencoder (DDAE) and deep neural network (DNN)-based SE models have also been proposed and extensively investigated [11-13]. In addition, to model the local temporal-spectral structures of a spectrogram efficiently, convolutional neural networks (CNN) have also been employed to further improve the SE performance [14, 15]. Most of these denoising models focus only on processing the magnitude spectrogram (e.g., log-power spectra (LPS)) and leave the phase in its original noisy form. This may be because no clear structure exists in a phase spectrogram, precisely estimating clean phases from noisy counterparts [16] is difficult.

Several recent studies have revealed the importance of phase when spectrograms are resynthesized back into time-domain waveforms [17, 18]. For example, Paliwal *et al.* confirmed the importance of phase for perceptual quality in speech enhancement, especially when window overlap and length of the Fourier transform increase [17]. To further improve the performance of speech enhancement, phase information is considered in some up-to-date studies [16, 19, 20]. Williamson *et al.* [16, 19] employed a DNN to estimate the complex ratio mask (cRM) from a set of complementary features, and then the magnitude and phase can be jointly enhanced through cRM. Although having been confirmed to provide satisfactory denoising performance, these methods still need to map features between time and frequency domains for analysis and resynthesizing through the (inverse) Fourier transform.

In the field of ASR, several studies have shown that deep-learning-based models with raw waveform inputs can achieve higher accuracy than those with hand-crafted features (e.g., MFCC) [21-26]. Because the acoustic patterns in time domain can appear in any positions, most of these methods employ CNN to detect useful information efficiently. However, in the field of speech enhancement, directly using the raw waveforms as system inputs has not been well studied. When compared to ASR, in addition to distinguishing speech patterns from noise, SE must further generate the enhanced speech outputs. In the time domain, each estimated sample point has to cooperate with its neighbors to represent frequency components. This interdependency may produce a laborious model in generating high and low frequency components simultaneously. Until recently, wavenet [27] was proposed and successful models raw audio waveforms through sample wise prediction and dilated convolution.

In this study, we investigate the capability of different deep-learning-based SE methods with raw waveform features. We first note that the fully connected layers may not well preserve local information to generate high frequency components. Therefore, we employ a fully convolutional network (FCN) model to enable each output sample to depend locally on the neighboring input regions. FCN is very similar to a conventional CNN except that the top fully connected layers are removed [28]. Recently, FCN has been proposed for SE [29] to process the magnitude spectrum. In addition, since the effect of convolving a time domain signal *x*(*t*) with a filter *h*(*t*) equals to multiplying its frequency representation *X*(*f*) with the frequency response of the filter *H*(*f*) [30]. Hence, it may be unnecessary to explicitly mapping waveform to spectrogram for speech enhancement. Based on the unique properties of FCN and the successful results in [29], we adopted FCN to construct our waveform-in and waveform-out system. Experimental results show that compared to DNN and CNN, the proposed FCN model can not only effectively recover the waveform but also dramatically reduce the number of parameters.

## II. RAW WAVEFORM SPEECH ENHANCEMENT

The goal of SE is to improve the intelligibility and quality of a noisy speech signal [31]. Because the properties in the log domain are more consistent with the human auditory system, conventionally, the log power spectrum is extracted from a raw speech signal for deep-learning-based denoising models [12, 13, 32-34]. However, employing LPS as features produces two drawbacks. First, phase components have not been well considered in LPS. In other words, when the enhanced speech signal is synthesized back to the time domain, the phase components are simply borrowed from the original noisy speech, which may degrade the perceptual quality of enhanced speech [17, 18]. Second, the (inverse) Fourier transform must be applied for mapping between time and frequency domains, thus increasing the computation load. In this study, we propose raw waveform-based SE system as illustrated in Fig.1 and explore solutions to address these issues.

### A. Characteristics of Raw Waveform

The characteristics of a signal represented in the time domain are very different from those in the frequency domain. In the frequency domain, the value of a feature (frequency bin) represents the energy of the corresponding frequency component. However, in the time domain, a feature (sample point) alone does not carry much information; it must combine information from its neighbors in order to represent a certain frequency component. For example, a sample point must be very different or very similar to its neighbors to represent high or low frequency components, respectively. This interdependency may produce a laborious model for representing high and low frequency components simultaneously. It may also cause many denoising models to choose to work in the frequency domain rather than in the time domain [7-10, 12]. In addition, unlike the spectrogram of speech signal (e.g., the consonants usually occupy only high frequency bins, whereas the repeated patterns of formants usually concentrate on low-to-middle frequency bins), the patterns in the time domain can appear in any position. This suggests that the convolution operation can efficiently find useful locally acoustic information. Therefore, most studies have employed the CNN model for analyzing raw waveforms [21-25, 27].

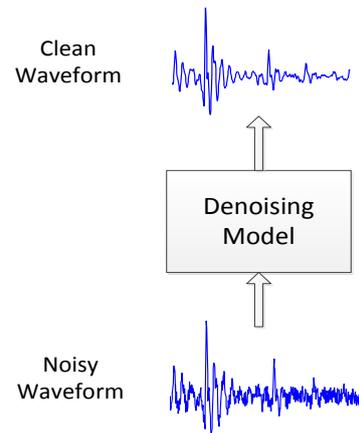

Fig. 1. Speech enhancement using raw waveform.

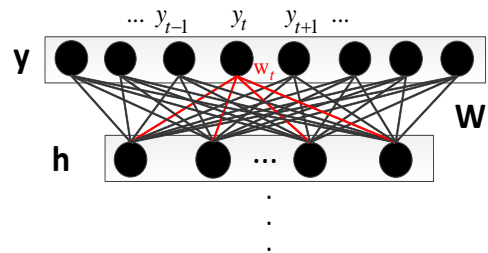

Fig. 2. Relation between output layer and last hidden layer in a fully connected layer.

### B. Problems in Fully Connected Layers for Modeling Raw Waveform

Using artificial neural networks (ANNs) for waveform-based speech enhancement can date back as early as to 1980's. In [35, 36], Tamura and Waibel used an ANN to predict short window of clean speech waveforms from noisy ones. They found that the ANN-enhanced waveform has no higher formant structures and gave some explanations by analyzing the weight matrix between last hidden layer and output layer. This phenomenon is also observed in our DNN and CNN-enhanced waveform.

The output layer and last hidden layer in DNN and CNN are linked in a fully connected manner, as shown in Fig. 2. We argue that this kind of connection produces difficulties in modeling high and low frequency components of waveform simultaneously. The relation between the output and last hidden layers can be represented by the following equation (bias is neglected here for simplicity).

$$\mathbf{y} = \mathbf{Wh} \qquad (1)$$

where $\mathbf{y} = [y_1 \dots y_t \dots y_N]^T \in R^{N \times 1}$ denotes the output sample points of the estimated waveform, and *N* is the number of points in a frame. $\mathbf{W} = [\mathbf{w}_1 \dots \mathbf{w}_t \dots \mathbf{w}_N]^T \in R^{N \times h}$ is the weight matrix, *h* is the number of nodes in the last hidden layer, and $\mathbf{w}_n \in R^{h \times 1}$ is the weight vector that connects the hidden layer $\mathbf{h} \in R^{h\times 1}$ and the output sample $y_n$. In other words, each sample point can be represented as:

$$y_t = \mathbf{w}_t^T \mathbf{h} \quad (2)$$

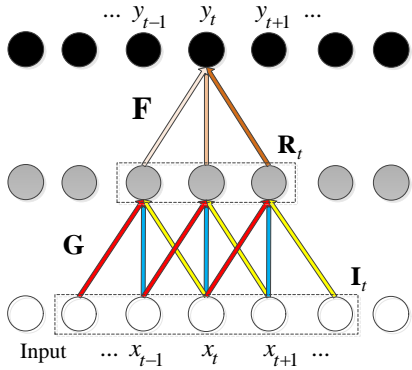

Fig. 3. Local connection in fully convolutional networks.

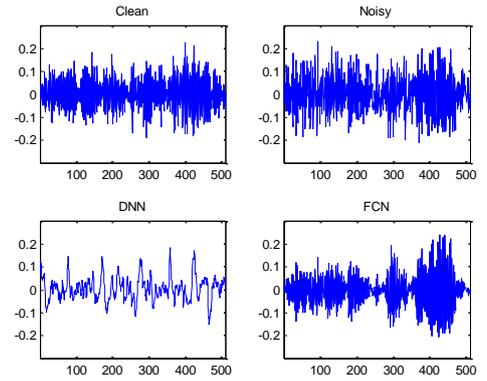

Fig. 4. Example of generating a high frequency signal by DNN and FCN.

With fixed $\mathbf{h}$, we consider two situations: 1) when $y_t$ is in the high frequency region, its value should be very different from its neighbors (e.g., $y_{t-1}$, $y_{t+1}$), which implies that $\mathbf{w}_t$ and $(\mathbf{w}_{t-1}, \mathbf{w}_{t+1})$ cannot be highly correlated; 2) when $y_t$ is in the low frequency region, we can deduce that $\mathbf{w}_t$ and $(\mathbf{w}_{t-1}, \mathbf{w}_{t+1})$ should correlate. However, because $\mathbf{W}$ is fixed after training, situations 1) and 2) cannot be satisfied simultaneously. Therefore, it is difficult to "learn" the weights in fully connected layers to generate high and low frequency parts of a waveform simultaneously. Please note that here we double quotes the term to emphasize that this structure only makes learning more difficult, not implying DNN cannot represent this mapping function. In fact, from universal approximation theorem [37], a DNN can approximate any memory-less function when given appropriate parameters; however, it does not guarantee those parameters can be learned.

In fact, the hidden fully connected layers also encounter difficulties modeling raw waveforms. We discuss this problem in greater detail in Section V.

### III. FCN

In the previous section, we showed that fully connected layers may not model raw waveforms precisely. Therefore, in this study, we try to apply FCNs, which do not contain any fully connected layers, to perform SE in the waveform domain. FCN is very similar to the conventional CNN, except that all the fully connected layers are removed. This can produce several benefits and has achieved great success in the field of computer vision for modeling raw pixel outputs [28]. The advantage of discarding fully connected layers is that the number of parameters in the network can be dramatically reduced, thus making FCNs particularly suitable for implementations in mobile devices with limited storage capacity. In addition, each output sample in FCN depends only locally on the neighboring input regions as shown in Fig. 3. This is different from fully connected layers in which the local information and the spatial arrangement of the previous features cannot be well preserved.

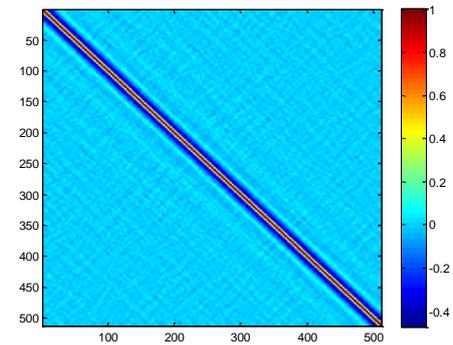

Fig. 5. Correlation matrix of the last weight matrix $\mathbf{W}$ in DNN.

More specifically, to explain why FCN can model high and low frequency components of raw waveforms simultaneously, we start with the connections between the output and last hidden layers. The relation between output sample $y_t$ and the connected hidden nodes $\mathbf{R}_t$ (also called receptive field) can be simply represented by the following equation (bias is neglected for simplicity).

$$y_t = \mathbf{F}^T \mathbf{R}_t \quad (3)$$

where $\mathbf{F} \in R^{f\times 1}$ denotes one of the learned filters, and $f$ is the size of the filter. Please note that $\mathbf{F}$ is shared in the convolution operation and is fixed for every output sample. Therefore, if $y_t$ is in the high frequency region, $\mathbf{R}_t$ and $(\mathbf{R}_{t-1}, \mathbf{R}_{t+1})$ should not be very similar. Whether $\mathbf{R}_t$ is different from its neighbors depends on the filtered outputs of previous locally connected nodes (input) $\mathbf{I}_t$. For example, when the input $\mathbf{I}_t$ is in the high frequency region, and the filter $\mathbf{G}$ is a high-pass filter, then $\mathbf{R}_t$ (and hence $y_t$) may also be extremely different from its neighbors. This argument can also hold for the low frequency case. Therefore, FCN can well preserve the local input information and handle the difficulty of using fully connected layers to model high and low frequency components simultaneously. When comparing the locations of subscript $t$ from (2) to (3), it can be observed that $t$ changes from the model ($\mathbf{w}_t$) to connected nodes ($\mathbf{R}_t$). This implies that in the fully connected case, the model has to deal with the interdependency between output samples, whereas in FCN, the connected nodes handle the interdependency.

TABLE I
PERFORMANCE COMPARISON OF DIFFERENT MODELS WITH RESPECT TO STOI AND PESQ.

| | DNN-baseline (LPS) | | DNN (waveform) | | CNN (waveform) | | FCN (waveform) | |
|---|---|---|---|---|---|---|---|---|
| SNR (dB) | STOI | PESQ | STOI | PESQ | STOI | PESQ | STOI | PESQ |
| 12 | 0.814 | 2.334 | 0.737 | 2.548 | 0.788 | 2.470 | 0.874 | 2.718 |
| 6 | 0.778 | 2.140 | 0.715 | 2.396 | 0.753 | 2.302 | 0.833 | 2.346 |
| 0 | 0.717 | 1.866 | 0.655 | 2.118 | 0.673 | 2.011 | 0.758 | 1.995 |
| -6 | 0.626 | 1.609 | 0.549 | 1.816 | 0.561 | 1.707 | 0.639 | 1.719 |
| -12 | 0.521 | 1.447 | 0.429 | 1.573 | 0.441 | 1.453 | 0.506 | 1.535 |
| Avg. | 0.691 | 1.879 | 0.617 | **2.090** | 0.643 | 1.989 | **0.722** | 2.063 |

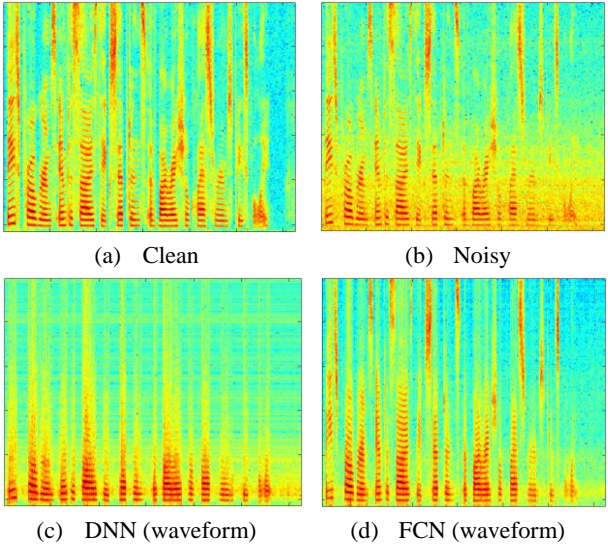

Fig. 6. Spectrograms of a TIMIT utterance: (a) clean speech, (b) noisy speech (engine noise), (c) enhanced waveform by DNN, and (d) enhanced waveform by FCN.

(a) Clean  (b) Noisy  (c) DNN (waveform)  (d) FCN (waveform)

## IV. EXPERIMENTS

### A. Experimental Setup

In our experiments, the TIMIT corpus [38] was used to prepare the training and test sets. For the training set, 600 utterances were randomly selected and corrupted with five noise types (Babble, Car, Jackhammer, Pink, and Street) at five SNR levels (-10 dB, -5 dB, 0 dB, 5 dB, and 10 dB). For the test set, we randomly selected another 100 utterances (different from those used in the training set). To make experimental conditions more realistic, both noise types and SNR levels of the training and test sets were mismatched. Thus, we adopted three other noise signals: white Gaussian noise (WGN), which is a stationary noise; and an engine noise and a baby cry, which are two non-stationary noises, using another five SNR levels (-12 dB, -6 dB, 0 dB, 6 dB, and 12 dB) to form the test set. All the results reported in Section IV-B were averaged across the three noise types.

In this study, 512 sample points were extracted from the waveforms to form a frame for the proposed SE model. In addition, the 257 dimensional LPS vectors were also obtained from the frames for the baseline system. The CNN in this experiment had four convolutional layers with padding (each layer consisted of 15 filters each with a filter size of 11) and two fully connected layers (each with 1024 nodes). FCN had the same structure as that of CNN, except the fully connected layers were each replaced with another convolutional layer. DNN had only four hidden layers (each layer consisting of 1024 nodes), because when it grows deeper, the performance starts to saturate as a result of the optimization issue [39]. All the models employ parametric rectified linear units (PReLUs) [40] as activation functions and are trained using Adam [41] with batch normalization [42] to minimize the mean square error between clean and enhanced waveform.

To evaluate the performance of the proposed models, the perceptual evaluation of speech quality (PESQ) [43] and the short-time objective intelligibility (STOI) scores [44] were used to evaluate the speech quality and intelligibility, respectively.

### B. Experimental Results

*1) Qualitative Comparison*: In this section, we investigate different deep learning models for SE with raw waveform. Fig. 4 shows an example of modeling a high frequency signal by DNN and FCN. In this figure, we can observe that for DNN to produce the corresponding high frequency signal as FCN is difficult. The same phenomenon can also be observed in CNN (not shown because of space restrictions). As mentioned in Section II-B, the failing of modeling high-frequency components is due to the natural limitation of fully connected layers. More specifically, since the high frequency components in speech are rare, this generally causes DNN and CNN to sacrifice the high frequency components in the optimization process. To further verify this claim, the correlation matrix $\mathbf{C}$ of the last weight matrix $\mathbf{W}$ in DNN is presented in Fig. 5. The element of $\mathbf{C}$ is defined as follows:

$$C_{ij} = \frac{(\mathbf{w}_i - \overline{\mathbf{w}_i})^T (\mathbf{w}_j - \overline{\mathbf{w}_j})}{\|\mathbf{w}_i - \overline{\mathbf{w}_i}\|_2 \|\mathbf{w}_j - \overline{\mathbf{w}_j}\|_2} \quad \forall\ 1 \leq i, j \leq 512 \quad (4)$$

here, $\mathbf{w}_i \in R^{1024 \times 1}$ is the weight vector, and $\overline{\mathbf{w}_i}$ is the mean of $\mathbf{w}_i$. The diagonal regions of $\mathbf{C}$ show that each weight vector is related only to its neighboring vectors and that the correlation drops to zero when the two vectors are a considerable distance from each other. In addition, the correlation coefficient of two neighboring vectors approximately reaches 0.9, implying that the generated samples strongly correlate. This explains why for DNN (and CNN) to generate high frequency waveform is arduous.

We next present the following: the spectrograms of a clean speech utterance, the same utterance corrupted by the engine noise, DNN-enhanced waveform, and FCN-enhanced waveform in Fig. 6(a), (b), (c), and (d), respectively. When comparing Fig. 6(a) and (c), we can clearly observe that the high frequency components of speech are missing in the spectrogram of DNN-enhanced waveform. This phenomenon can also be observed in CNN (not shown because of space restrictions) but is not as serious as in the DNN case. However, by comparing Fig. 6(a) and (d), we can note that speech components are well preserved and noise is effectively removed.

*2) Quantitative Comparison*: Finally, Table I presents the results of the average STOI and PESQ scores on the test set, based on different models and features. From this table, we can see that the waveform-based DNN achieved the highest PESQ score and the worst STOI score, suggesting that a good balance cannot be achieved between the two goals of speech enhancement (improving both the intelligibility and quality of a noisy speech signal). This may be because the model eliminates too many speech components when removing noise. By contrast, FCN can achieve the highest STOI score and a satisfactory PESQ score. It is worth nothing that because the fully connected layers were removed, the number of weights involved in FCN was approximately only 0.2% when compared to that involved in DNN and CNN.

## V. DISCUSSION

We also noted that the issue of missing high frequency components becomes critical when the number of fully connected layers increases. This implies that the hidden fully connected layers actually also have difficulties in modeling waveform. The reason may be that it is crucial to preserve the relations between sample points in time domain to represent a certain frequency component. However the mapped features by the fully connected layer are abstract and do not retain the spatial arrangement of the previous features. In other words, fully connected layers destroy the correlation between features, making it difficult to generate waveforms. This effectively explains why CNN has relatively minor problems regarding missing high frequency components when compared to DNN, because CNN contains fewer fully connected layers.

The generation of high frequency components by DNN is also influenced by how the data is fed in. In general, waveform is presented to DNN by sliding the input window across the noisy speech. At each step, the window is shifted by an increment, $L$, between 1 and the window length $M$ (512 in our experiment). When $L = M$, the estimation window moves along without overlap and this setting was adopted in previous section. We found that in the case of a single time step increment, $L = 1$, which most closely corresponds to filter implementations [45], the high frequency components can be successfully generated as FCN. Fig. 7 illustrates the output frames with window shift increment 1 and 512 and the enhanced waveform when the clean speech is in low and high frequency cases, respectively. It can be observed that when the shift increment is 1, DNN can successfully generated high frequency components. Note that the DNN used in these two settings is the same; the only difference is how the data is given to the model. In fact, when $L = 1$, we can treat the whole DNN as a filter in the convolution, and the relation between output and last hidden layer is similar to FCN. Specifically, if we take the first node of output layer in DNN as estimated output (as in Fig. 7), then every output sample is generated through fixed weights $\mathbf{w}_1$, which are similar to the role of learned filters $\mathbf{F}$ in (3).

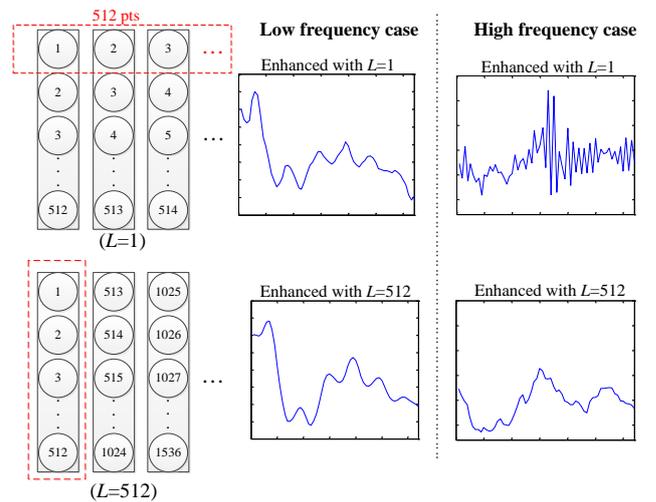

Fig. 7. Frames with window shift increment 1 and 512.

From this discussion, we can conclude that since the weight vectors in last fully connected layer are highly correlated to each other, it is difficult for them to produce high frequency waveform (as in the lower part of Fig. 7). However, if we only use one node, then the problem can be solved (as in the upper part of Fig. 7). Because in this case, each estimated sample point is decided by fixed weights and different inputs rather than fixed input and different weights as in the $L = 512$ case. Although applying DNN in a filter way ($L = 1$) can solve the missing high frequency problem, it is very inefficient compared to FCN.

## VI. CONCLUSIONS

The contribution of our study is two-fold. First, we investigated the capability of different deep-learning-based SE methods with raw waveform inputs. The results indicated that fully connected layers may not be necessary because: 1) they dramatically increase the number of model parameters; 2) they have limited capability to preserve the correlation between features, which is very important for generating waveforms. Second, to overcome this problem, we employed FCN in our study and confirmed that it yields better results compared to those of DNN with LPS inputs. In the future, to enhance (optimize) each utterance as a whole, we will apply FCN in an utterance-based manner instead of frame-wise processing.